\documentclass[10pt,twocolumn,letterpaper]{article}
\usepackage{cvpr}
\usepackage{times}
\usepackage{epsfig}
\usepackage{graphicx}
\usepackage{amsmath}
\usepackage{amssymb}
\usepackage{caption}
\usepackage{multicol}
\usepackage{tabularx}
\usepackage{booktabs}

\usepackage[pagebackref=true,breaklinks=true,letterpaper=true,colorlinks,bookmarks=false]{hyperref}

\cvprfinalcopy 
\def\cvprPaperID{****}

\ifcvprfinal\fi
\begin{document}
\setcounter{topnumber}{2}

\title{M2U-Net: Effective and Efficient Retinal Vessel Segmentation for Real-World Applications}

\author{Tim Laibacher\\
City, University of London\\
{\tt\small tim.laibacher@city.ac.uk}
\and 
\and
Tillman Weyde\\
City, University of London\\
{\tt\small t.e.weyde@city.ac.uk}
\and
Sepehr Jalali\\
University College London\\
{\tt\small s.jalali@ucl.ac.uk}
}

\maketitle
\begin{abstract}
In this paper, we present a novel neural network architecture for retinal vessel segmentation that improves over the state of the art on two benchmark datasets, is the first to run in real time on high resolution images, and its small memory and processing requirements make it deployable in mobile and embedded systems. 

The M2U-Net has a new encoder-decoder architecture that is inspired by the U-Net. It adds pretrained components of MobileNetV2 in the encoder part and novel contractive bottleneck blocks in the decoder part that, combined with bilinear upsampling, drastically reduce the parameter count to 0.55M compared to 31.03M in the original U-Net. 

We have evaluated its performance against a wide body of previously published results on three public datasets. 
On two of them, the M2U-Net achieves new state-of-the-art performance by a considerable margin.
When implemented on a GPU, our method is the first to achieve real-time inference speeds on high-resolution fundus images.
We also implemented our proposed network on an ARM-based embedded system where it segments images in between 0.6 and 15 sec, depending on the resolution. 
Thus, the M2U-Net enables a number of applications of retinal vessel structure extraction, such as early diagnosis of eye diseases, retinal biometric authentication systems, and robot assisted microsurgery.
\end{abstract}

\section{Introduction}
The retinal vasculature in each human eye is unique and, in the absence of pathologies, stays unaltered throughout the lifetime~\cite{ortega_personal_2009,simon_new_1935}. 
This trait makes the automatic segmentation of the vessel structure suitable for two main applications: 

First, in the early detection of diseases that affect the vessel structure, such as diabetic retinopathy and wet age-related macular degeneration. 
Diabetic retinopathy can cause the growth of new blood vessels. 
Wet macular degeneration can result in atherosclerosis, which can lead to the narrowing of blood vessels and can affect the arteries to vein ratio ~\cite{abramoff_retinal_2010,hubbard_methods_1999}. 

Second, in biometric authentication systems, where the vessel structure is commonly used as a feature pattern by itself ~\cite{figueiredo_automated_2016,marino_personal_2006} or as a preprocessing step to extract feature points based on landmarks like bifurcations and crossovers of retinal vessels~\cite{ahmed_retina_2014,akram_retinal_2011,ortega_personal_2009}.
To achieve retinal identification, good segmentation of the principal vessels of each individual is  necessary~\cite{figueiredo_automated_2016}.
An emerging third application is in robotic intraocular microsurgery, where the instruments location relative to the retinal vessel structure is tracked in real time as proposed by Braun \etal~\cite{braun_eyeslam:_2018}.

For real-world deployments of vessel segmentation methods, not only the segmentation quality, but also model size and computational requirements in terms of memory and processing power are important factors. 
In a real-world retinal authentication system for example, it would be infeasible to deploy expensive and power-hungry server grade GPUs, often consuming more than 200W, at every point of authentication. 
Additionally, the usage of cloud-computing resources in such a setting is prohibitive due to higher latency and privacy concerns, arising from the processing of large amounts of sensitive personal biometric data in the form of retina fundus images. 
By deriving feature patterns of the retina vasculature directly at the point of authentication, without the actual fundus images leaving the device, the risk of leaking or exposing this sensitive information can be reduced.

In this work we propose a novel neural network architecture, called M2U-Net, for retinal vasculature segmentation. M2U-Net reaches state of the art quality and is small and fast enough for use in embedded and mobile environments, enabling use-cases in biometrics and mobile diagnosis.
Our specific contributions are as follows:
\begin{itemize}
    \item The M2U-Net architecture, which builds on pretrained components of MobileNetV2~\cite{sandler_inverted_2018}, is inspired by the U-Net architecture~\cite{ronneberger_u-net:_2015} and introduces contracting bottleneck blocks. 
    \item The proposed network is tested on three publicly available retina fundus image datasets with annotated vessel ground truth labels: 
    DRIVE~\cite{staal_ridge-based_2004}, CHASE\_DB1~\cite{fraz_ensemble_2012} and HRF~\cite{kohler_automatic_2013}.
    \item We demonstrate that M2U-Net achieves new state-of-the-art results on HRF and CHASE\_DB1 and new state of the art for mobile/embedded implementations on DRIVE. 
    Our work is the first to reach super-human performance on CHASE\_DB1 and real-time inference speeds on HRF and CHASE\_DB1. 
\end{itemize}

The remainder of this paper is structured as follows: 
Section 2 reviews related work with a focus on deep neural networks and efficient models. 
Section 3 introduces the M2U-Net architecture. 
Section 4 describes the implementation details and datasets.
Section 5 describes the results together with an additional ablation study.
Section 6 provides the conclusion with further research avenues. 

\section{Related work}
The task of retinal blood vessel segmentation falls into the computer vision subcategory of semantic segmentation, which in recent years has seen tremendous improvements in performance thanks to the introduction of novel deep neural network architectures~\cite{long_fully_2015,romera_erfnet:_2018,ronneberger_u-net:_2015}. 
Similarly, recent state-of-the-art methods in retinal blood vessel segmentation that focus on segmentation quality are dominated by various variations of deep neural networks~\cite{liskowski_segmenting_2016,maninis_deep_2016,yan_joint_2018}. 

Liskowski and Krawiec~\cite{liskowski_segmenting_2016} propose a patch-based approach, where a network that consists of a stack of convolutional layers followed by three fully-connected layers is trained with small patches of the input fundus image. %
In contrast, Maninis \etal~\cite{maninis_deep_2016} and Yan \etal~\cite{yan_joint_2018} train their networks with complete fundus images. 

Maninis \etal~\cite{maninis_deep_2016} extract intermediate feature maps of a VGG-16 network~\cite{simonyan_very_2014}, pretrained on ImageNet, which are upsampled via transposed convolutions and concatenated before applying a final $1 \times 1$ convolutional layer. 
Their method, called DRIU, achieves a state-of-the-art Dice score of 0.822 on DRIVE. %
Yan \etal~\cite{yan_joint_2018} train the U-Net model~\cite{ronneberger_u-net:_2015} with a joint-loss by appending two separate branches, one with a pixel-wise and one with a segment-level loss, that are trained simultaneously.

While these supervised methods accomplish good segmentation results, their computational requirements are substantial and as a consequence are commonly implemented on high-performance server-grade GPUs such as the NVIDIA TITAN class GPUs in \cite{liskowski_segmenting_2016,maninis_deep_2016,yan_joint_2018}. 

Additionally they either fail to reach the performance of unsupervised methods on very high-resolution datasets or, as a result of their computational requirements, can only be trained with small patches of the complete input image that fit into memory and thereby further increase the time it takes to segment the complete fundus image. 
For example, on the high resolution HRF dataset, the unsupervised method introduced by Annunziata \etal~\cite{annunziata_leveraging_2016} achieves a state-of-the art Dice score of 0.7578, while the best-performing supervised method~\cite{yan_joint_2018} achieves a Dice score of 0.7212.

Furthermore unsupervised methods are still dominant in works that focus on embedded or mobile systems and on execution speed ~\cite{arguello_gpu-based_2014,bendaoudi_flexible_2016,bibiloni_real-time_2018,koukounis_high_2014}. 
These methods often rely on matched filtering, contour tracing and morphological transformation techniques. 
In~\cite{arguello_gpu-based_2014}, a combination of matched filtering and contour tracing is proposed and implemented on GPU, realizing an execution time of 10ms on DRIVE. 
FPGA implementations of matched filtering based methods are introduced in \cite{bendaoudi_flexible_2016} and \cite{koukounis_high_2014} that accomplish execution times of 2ms and 52.3 ms respectively. %
In addition to morphological transformations, Bibiloni \etal~\cite{bibiloni_real-time_2018} use CLAHE and hysteresis thresholding to segment the retinal vasculature and report single-core execution speeds of 37ms on DRIVE, using a Intel i5-3340 CPU.%

To the best of our knowledge, \cite{xu_smartphone-based_2016} is the only existing work that has been implemented and tested on a mobile device, a Samsung Galaxy S5. 
The algorithm they introduced is based on visual saliency and incorporates orientation, morphological, spectral and intensity features and has an execution time of 118s on DRIVE images.

\section{Model architecture}

In this work we address this gap between unsupervised and supervised methods in retinal vessel segmentation by introducing a light-weight supervised method that can be deployed in embedded systems, can be trained on very high resolution images and produces competitive segmentation quality. 

\begin{figure}
\begin{center}
\includegraphics[width=0.48\textwidth]{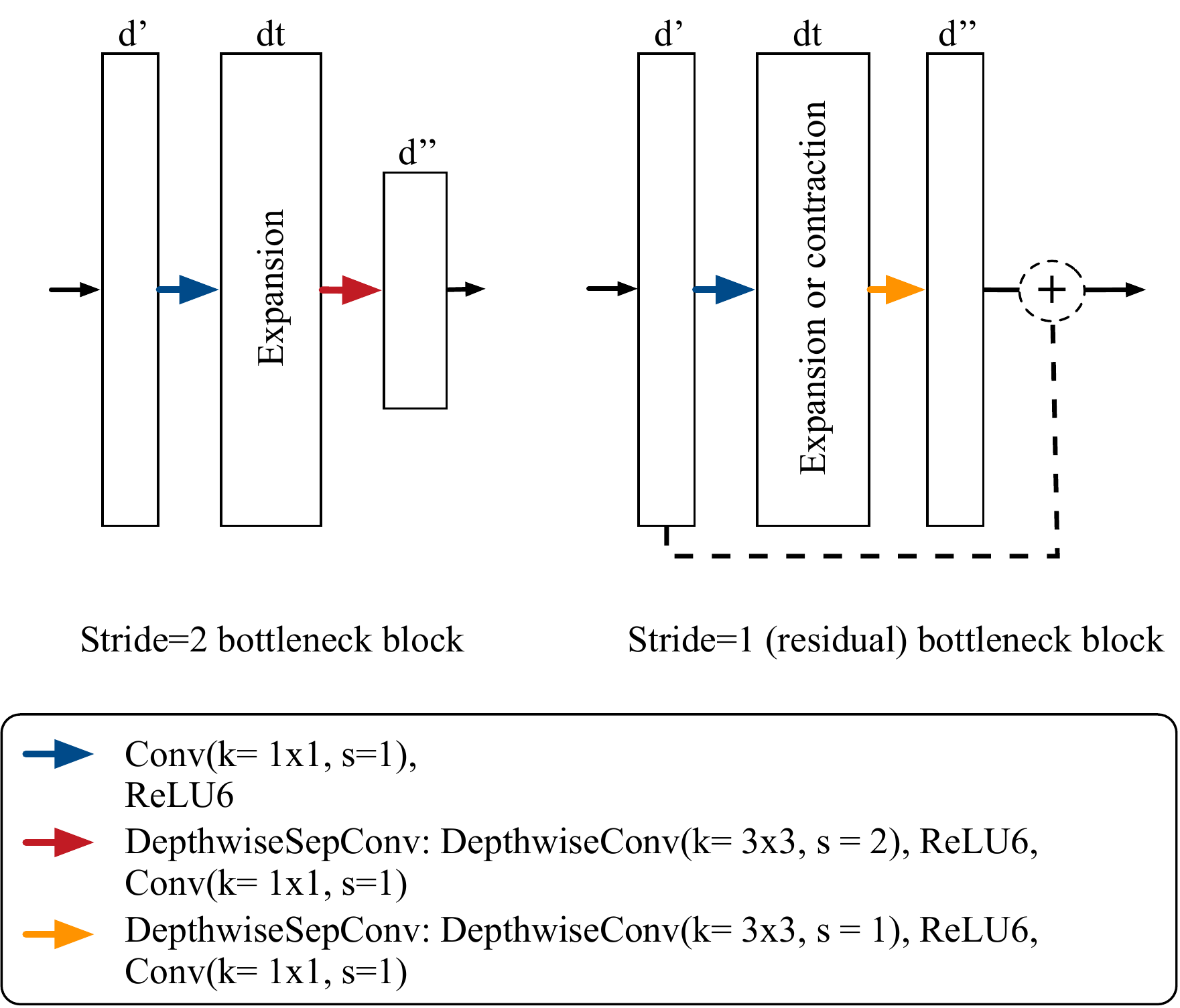}
\end{center}
   \caption{Bottleneck blocks: each box represents a multi-channel feature map. 
   	The height corresponds to the resolution, the width to the number of channels. 
   	Best viewed in color.}
\label{fig:blocks}
\end{figure}

Like the U-Net architecture, our proposed network can be conceptually divided into two parts: a contracting encoder part and an expanding decoder part.
For the encoder part, we employ the first fourteen layers of MobileNetV2 that were pretrained on ImageNet. The key building blocks, introduced by
Sandler \etal~\cite{sandler_inverted_2018} in the MobileNetV2 network, are inverted bottleneck blocks with Stride=1 and Stride=2, which are illustrated in Figure~\ref{fig:blocks}. 
Bottlenecks blocks with Stride=1 contain a residual connection if the number of input channels $d'$ is equal to the number of output channels $d''$. 
Bottleneck blocks with Stride=2 do not have a residual connection. 
Both types of blocks start with a $1\times1 $ convolution that expands the number of channels $d$ by a factor $t$, followed by a depthwise separable convolution~\cite{sifre_rigid-motion_2014} that consists of a depthwise convolution and a pointwise $1 \times 1$ convolution.

\begin{figure*}
\begin{center}
\includegraphics[width=1\textwidth]{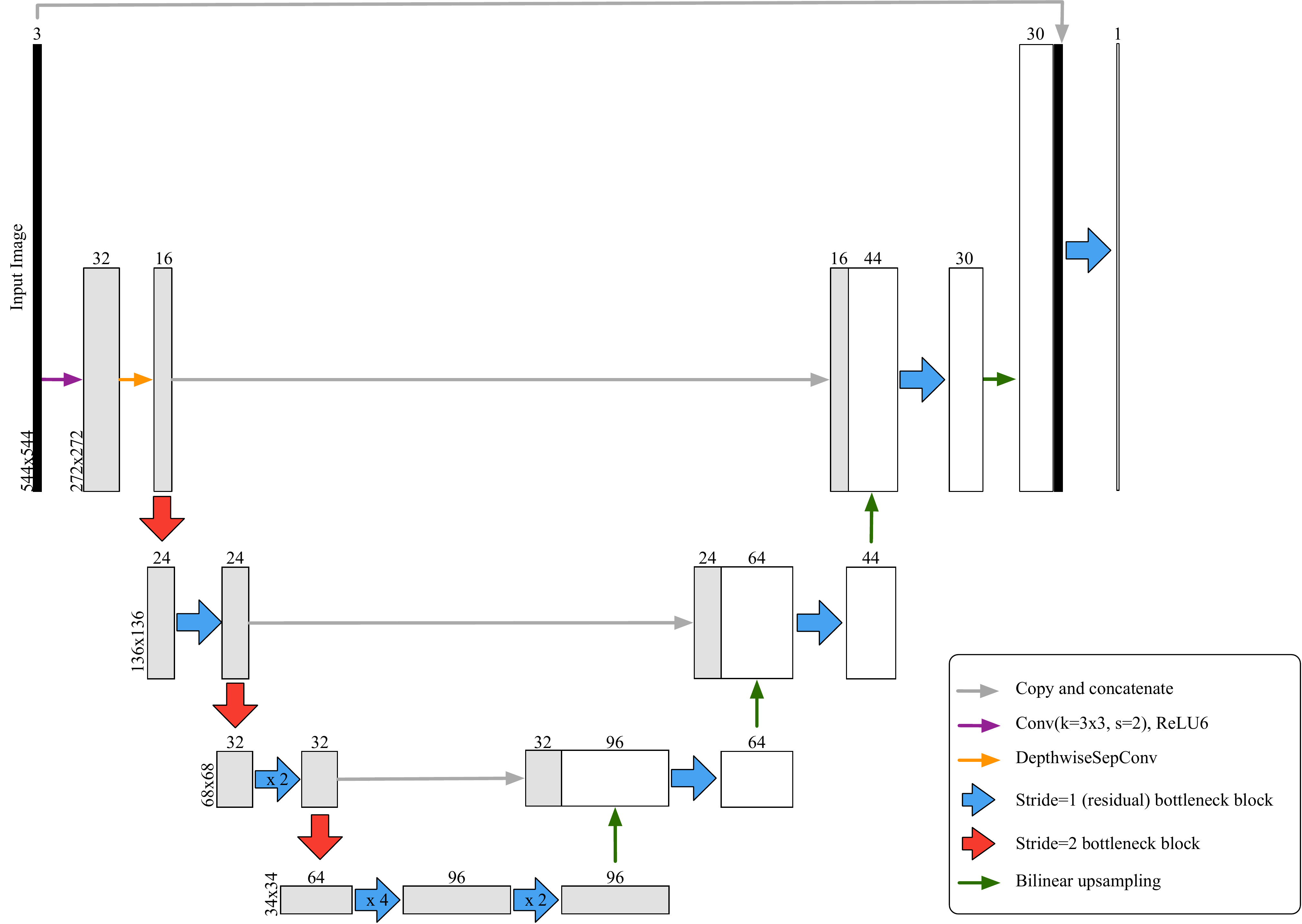}
\end{center}
  \caption{M2U-Net: Example for an input image of dimension $3 \times 544 \times 544$. 
  	Each white and grey rectangular box represents a multi-channel feature map. 
  	For illustration purposes the height corresponds to the resolution and the width to the number of channels. 
  	The encoder part feature maps are highlighted in grey. 
  	Best viewed in color.} 
\label{fig:mobile2unetdiagram}
\end{figure*}

In the decoder part we make use of the same Stride=1 bottleneck blocks as used in the encoder, but this time with a contracting factor of $t=0.15$ instead of using an expansion factor of $t=6$. These contracting bottleneck blocks do not contain a residual connection.
The contracting bottleneck blocks allow us to further reduce the amount of trainable parameters. 
Additionally we follow the recommendation of Fauw \etal~\cite{fauw_clinically_2018} and use parameter-free bilinear upsampling instead of transposed convolutional operations to upsample the spatial resolution of feature maps.
Table~\ref{tab:mobile2unet} and Figure~\ref{fig:mobile2unetdiagram} describe and illustrate the proposed architecture in more detail
\footnote{A PyTorch implementation is available at \url{https://github.com/laibe/M2U-Net}}.

\begin{table}[tbp]
\small
\centering
\begin{tabular}{@{}ccccccc@{}}
\toprule
Input                         & Operator                 & $t$  & $c$ & $n$ & $s$ & Params   \\ \midrule
$ 3 \times 544^2 $   & conv                     & -    & 32  & 1   & 2   & 928       \\
$ 32 \times 272^2 $  & dwisesep                 & 1    & 16  & 1   & 1   & 896       \\
$ 16 \times 272^2 $  & bottleneck               & 6    & 24  & 1   & 2   & 5,136     \\
$ 16 \times 136^2 $  & resbottleneck            & 6    & 24  & 1   & 1   & 8,832     \\
$ 24 \times 136^2 $  & bottleneck               & 6    & 32  & 1   & 2   & 10,000     \\
$ 32 \times 68^2 $    & resbottleneck           & 6    & 32  & 2   & 1   & 29,696    \\
$ 32 \times 68^2 $    & bottleneck              & 6    & 64  & 1   & 2   & 21,056    \\
$ 64 \times 34^2 $    & resbottleneck           & 6    & 64  & 3   & 1   & 162,816   \\
$ 64 \times 34^2$    & bottleneck               & 6    & 96  & 1   & 1   & 66,624    \\
$ 96 \times 34^2 $    & resbottleneck           & 6    & 96  & 2   & 1   & 236,544   \\
$ 96 \times 34^2 $     & upconcat               & -    & 128 & 1   & -   & -         \\
$ 128 \times 68^2 $    & bottleneck             & 0.15 & 64  & 1   & 1   & 4,023     \\
$ 64 \times 68^2 $    & upconcat                & -    & 88  & 1   & -   & -         \\
$ 88 \times 136^2 $  & bottleneck               & 0.15 & 44  & 1   & 1   & 1973      \\
$ 44 \times 136^2 $   & upconcat                & -    & 60  & 1   & -   & -         \\
$ 60 \times 272^2 $   & bottleneck              & 0.15 & 30  & 1   & 1   & 987       \\
$ 30 \times 272^2 $   & upconcat                & -    & 33  & 1   & -   & -         \\
$ 33 \times 544^2 $   & bottleneck              & 0.15 & 1   & 1   & 1   & 237       \\
$ 1 \times 544^2 $    & sigmoid                 & -    & 1   & 1   & -   & -         \\ \bottomrule
\end{tabular}
\caption{M2U-Net: Each row corresponds to an operation that uses stride $s$, expansion factor $t$, is repeated $n$ times and outputs $c$ channels. 
The encoder part consists of an initial convolution (\textbf{conv}) and a depthwise separable convolution (\textbf{dwisesep}) followed by a stack of \textbf{(residual) bottleneck} blocks. 
The encoder part consists of repeated instances of upsampling and concatenation operations (\textbf{upconcat}) followed by \textbf{bottleneck} operations. 
Encoder bottleneck blocks expand with a factor $t=6$, decoder bottleneck blocks contract with a factor $t=0.15$. 
Batch normalization is used after each convolutional operation.  
A final \textbf{sigmoid} layer converts the raw logits into probabilities.}
\label{tab:mobile2unet}
\end{table}

\section{Implementation details}
\subsection{Training setup and datasets} The proposed architecture is implemented in PyTorch. 
We use AdamW~\cite{loshchilov_fixing_2018} as optimization method with a learning rate of $0.001$.
A combined binary cross-entropy and Jaccard loss function $L_{JBCE}$ with a weighting factor $w=0.3$ as proposed in \cite{iglovikov_ternausnetv2:_2018} is utilized:
\begin{equation}\label{eq:lossjbce}
L_{JBCE} = L_{BCE} + w * (1-J),
\end{equation}
where $L_{BCE}$ is the binary cross entropy loss:
\begin{equation}
L_{BCE} = - \frac{1}{n} \sum^n_{i=1}(y_i log(\hat{y}_i) + (1-y_i) log(1-\hat{y}_i) )
\end{equation}
and $J$ the Jaccard index adapted for non-discrete objects:
\begin{equation}
J = \frac{1}{n} \sum^n_{i=1} \frac{y_i\hat{y}_i}{y_i+\hat{y}_i - y_i\hat{y}_i},
\end{equation}
$n$ denotes the number of pixels in a given image, the network's prediction probability output of the pixel belonging to the vessel class is denoted by $\hat{y}$ and the ground-truth by~$y$. 

For the DRIVE dataset we adopt the training-test split as proposed by the authors of the dataset (20 training and 20 test images)~\cite{staal_ridge-based_2004}. %
For CHASE\_DB1 we follow the suggestion of \cite{fraz_ensemble_2012}, where the first 8 images form the training set and the remaining 20 the test set. This split is also used in~\cite{orlando_discriminatively_2017}. 

For HRF we adopt the split as proposed by Orlando \etal~\cite{orlando_discriminatively_2017} and adapted in~\cite{yan_joint_2018}, who introduced the only supervised methods that have so far been evaluated on this dataset. 
The training set contains the first five images of each category (healthy, diabetic retinopathy and glaucoma); the remaining 30 images define the test set.

\begin{figure}
\begin{center}
\includegraphics[width=0.47\textwidth]{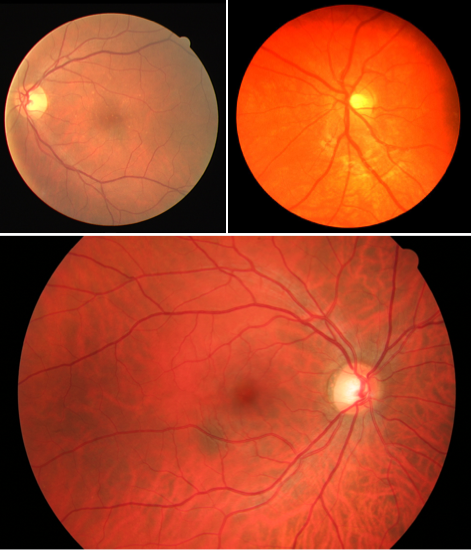}
\end{center}
   \caption{Examples of training images ($H\times W$). \textbf{Top left}: DRIVE ($584\times565$); \textbf{top right}: CHASE\_DB1 ($960\times999$); \textbf{bottom}: HRF ($2336\times3504$).}
\label{fig:trainingimg}
\end{figure}

Examples of training images of DRIVE, CHASE\_DB1 and HRF together with their resolution are shown in Figure~\ref{fig:trainingimg}. 
DRIVE and CHASE\_DB1 images contain large black boarders that do not contain any valuable information. 
To decrease the number of background pixels and to ensure that the input resolution is a multiple of 16, a requirement for M2U-Net, U-Net~\cite{ronneberger_u-net:_2015} and ERFNet~\cite{romera_erfnet:_2018}, we perform the following crops: 
\begin{itemize}
\item On DRIVE we take a $544 \times 544$ center crop. 
\item On CHASE\_DB1 we crop 18 pixels on the left and 21 on the right, resulting in a resolution of $960\times960$. 
\item On HRF we do not perform cropping as the original resolution of $2336 \times 3504$ is already a multiple of 16.
\end{itemize}

No other preprocessing is conducted and during training, a set of random augmentations are applied: rotations, horizontal and vertical flips, elastic distortions and changes in brightness, contrast, saturation and hue. 
Image augmentations are commonly used in biomedical image analysis tasks, especially when working with small datasets, as they can improve accuracy and generalization~\cite{ching_opportunities_2018}. 
We limit the rotations to a range from $-15$ to $+15$ degrees. Brightness, contrast, saturation and hue are changed by a small random factor in the range $\mathbb{R} \cap [1-c,1+c]$, where $c$ determines the freedom of change. The following values are selected: $c_{brightness} =0.3$, $c_{contrast}=0.3$, $c_{saturation}=0.02$ and $c_{hue}=0.02$.
Elastic transformations are governed by the grid size and the magnitude, for which we selected values of $8\times8$ and $1$ respectively~\cite{bloice_augmentor_2017}. 

Since our datasets contain a very small number of training images we derive separate validation sets from the training sets, using the same random augmentations.
During training, the model with the highest dice score on the validation set is selected. 
Training is stopped after 300 epochs. 

We compare the performance of M2U-Net versus the performance of ERFNet~\cite{romera_erfnet:_2018} and U-Net~\cite{ronneberger_u-net:_2015} that are trained with identical training parameters. 
ERFNet is a recent efficient real-time semantic segmentation method, originally applied on urban scene segmentation tasks. 
We deviate from the original U-Net implementation by using zero-padding to ensure the resolution of the final feature map is equal to the resolution of the input image. 
It should also be noted that in contrast to ERFNet and M2U-Net that utilize pretrained encoders, the U-Net was trained from scratch and acts as a baseline only.

On DRIVE we use a batch-size of 4 for ERFNet and M2U-Net and a batch size of 2 for U-Net due to it's higher memory footprint. 
On CHASE\_DB1 we use a batch-size of 2 for ERFNet and M2U-Net. 
The U-Net can not be trained on CHASE\_DB1 and HRF without reverting to a patch-based training approach, due to insufficient GPU memory in our training setup (11 GB). 
Similarly the memory requirement of ERFNet for HRF is too large for our setup. 
As a consequence we train only M2U-Net on HRF, using a batch-size of 1. 

\subsection{Evaluation metrics}
Since in our segmentation task the number of non-vessel pixels is far larger than the number of vessel pixels, accuracy is not a suitable evaluation measurement. 
Instead we follow Maninis \etal~\cite{maninis_deep_2016} and plot precision (Pr) and recall (Re) at various thresholds of the probability map output and compute the Dice score (also referred to as F1-score) as our key summary evaluation criteria. It is commonly expressed in terms of true positives ($TP$), false negatives ($FN$) and false positives ($FP$):
\begin{equation}
Dice = \frac{2* TP}{2*TP + FN + FP } = 2 \frac{Pr*Re}{Pr+Re}
\end{equation}

Another common evaluation measurement in semantic segmentation tasks, the Jaccard index, has a monotonically increasing relationship with the Dice score and as a consequence both measures are equivalent for comparison purposes \cite{pont-tuset_supervised_2016}. 
We report the Dice score, since it is more common in the literature on retinal vessel segmentation. 
For comparison purposes with methods that do not report Dice scores, we additionally report accuracy and area under the receiver operating characteristic curve (AuC). 
We adjust the number of true negatives by the number of pixels that were cropped to ensure a fair comparison of accuracy and AuC with previous works.

\section{Results}
\begin{table*}[tbp]
\small
\begin{center}
\begin{tabular}{@{}ccccccccccc@{}}
\toprule
                    &                 &                 & \multicolumn{4}{c}{DRIVE}                  & \multicolumn{4}{c}{CHASE\_DB1}                           \\ \cmidrule(lr){4-7} \cmidrule(lr){8-11}
Model               & Params          & Size            & Dice            & MAdds         & ARM              & ARM TVM & Dice          & MAdds          & ARM            & ARM TVM          \\ \midrule
2nd human annot.    & -               & -               & 0.7881          & -             & -                & -     & 0.7686          & -             &  -             & -              \\  
U-Net~\cite{ronneberger_u-net:_2015}  & 31.03M          & 119.0MB         & 0.7941          & 246.6B        & 141min           & 58.1s     & -               & -             & -              & -              \\
ERFNet~\cite{romera_erfnet:_2018}             & 2.06M           & 8.0MB           & 0.8022          & 2.8B          & 69.8s            & -     & 0.7994          & 51.3B         & 336s           & -              \\
\textbf{M2U-Net (ours)}            & \textbf{0.55M}  & \textbf{2.2MB}  & \textbf{0.8091}          & \textbf{1.4B} &\textbf{5.87s}    & \textbf{577ms} & \textbf{0.8006}          & \textbf{4.4B} & \textbf{23.5s} & 1.67s          \\ \bottomrule
\end{tabular}
\end{center}
\caption{Results on DRIVE with input dimension $3\times544\times544$ and CHASE\_DB1 with input dimension $3\times960\times960$. 
\textbf{Params}: Number of parameters. \textbf{Size}: Size of the weight file on disk. \textbf{Dice}: Dice score (= F1-Score). \textbf{MAdds}: Number of Multiply-Add operations. \textbf{ARM}: Rockchip RK3399; NNPACK single thread Cortex-A72 inference time.  \textbf{ARM TVM}: Rockchip RK3399; TVM dual Cortex-A72 inference time.}
\label{tab:resultsfourmodels1}
\end{table*}
\begin{table*}[tbp]
\small
\begin{center}
\begin{tabular}{@{}cccccccccc@{}}
\toprule
             &                                                                     & \multicolumn{4}{c}{DRIVE}                                          & \multicolumn{4}{c}{CHASE\_DB1}   \\ \cmidrule(lr){3-6} \cmidrule(lr){7-10}
Method                                              & Platform                     & Time         & Dice            & Acc             & AuC             & Time         & Dice            & Acc             & AuC            \\ \midrule
2nd human observer                                  & -                            & -            & 0.7881          & 0.9472          & -               & -            & 0.7686          & 0.9538          & -               \\
\textbf{Unsupervised}                               &                              &              &                 &                 &                 &              &                 &                 &                 \\ \hline
Arg{\"u}ello \etal~\cite{arguello_gpu-based_2014}   & NVIDIA GTX 680               & 10ms         & -               & 0.9431          & -               & -            & -               & -               & -               \\
Azzopardi \etal~\cite{azzopardi_trainable_2015}     & 2GHz CPU               & -            & -               & -               & -               & 25s          & -               & 0.9387          & 0.9487          \\
Bendaoudi \etal~\cite{bendaoudi_flexible_2016}      & Xilinx Kintex-7 FPGA\textsuperscript{*} & \textbf{2ms} & -    & 0.9218          & 0.9207          & -            & -               & -               & -               \\
Bibiloni \etal~\cite{bibiloni_real-time_2018}       & Intel Core i5-3340           & 37ms         & 0.7521          & 0.938           & -               & -            & -               & -               & -               \\
Jiang \etal~\cite{jiang_fast_2017}                  & Intel Core i7 Duo            & 1.677s       & -               & 0.9588          & -               & -            & -               & -               & -               \\
Koukounis \etal~\cite{koukounis_high_2014}          & Spartan 6 FPGA\textsuperscript{*} & 52.3ms  & -               & 0.9240          & 0.9008               & -            & -               & -               & -          \\
Xu \etal~\cite{xu_smartphone-based_2016}            & Samsung Galaxy S5\textsuperscript{*} & 118s & -               & 0.933           & 0.959           & -            & -               & -               & -  \\
Zhang \etal~\cite{zhang_robust_2016}                & 2.7GHz CPU                  & 20s            & -         & 0.9476               &  0.9636              & -            & -               & 0.9452          & 0.9487          \\
\textbf{Supervised}                                 &                              &              &                 &                 &                 &              &                 &                 &                 \\ \hline
Fraz \etal~\cite{fraz_ensemble_2012}                & Intel Core2Duo               & 100s         & 0.7929          & 0.9480          & 0.9747          & -            & 0.7566          & 0.9469          & 0.9712          \\
Fu \etal~\cite{fu_deepvessel:_2016}                 & -                            & 1.3s         & -               & 0.9523          & -               & -            & -               & -               & -               \\
Li \etal~\cite{li_cross-modality_2016}              & AMD Athlon II X4             & 70s          & -               & 0.9527          & 0.9738          & 70s          & -               & 0.9581          & 0.9716          \\
Liskowski \etal~\cite{liskowski_segmenting_2016}    & NVIDIA TITAN                 & -            & -               & 0.9535          & \textbf{0.9790} & -            & -               & -               & -               \\
Maninis \etal~\cite{maninis_deep_2016}              & NVIDIA TITAN-X               & 85ms         & \textbf{0.8220} & -               & -               & -            & -               & -               & -               \\
Marin \etal~\cite{marin_new_2011}                   & Intel Core2Duo               & 90s          & -               & 0.9452          & -               & -            & -               & -               & -               \\
Orlando \etal~\cite{orlando_discriminatively_2017}  & Intel Xeon E5-2690           & 1s           & 0.7857          & -               & 0.9507          & 2.7s         & 0.7332          & -               & -               \\
Roychowdhury \etal~\cite{roychowdhury_blood_2015}   & Intel Core i3                & 3.115s       & -               & 0.9520          & 0.9620          & 11.71s       & -               & 0.9530          & 0.9532          \\
Yan \etal~\cite{yan_joint_2018}                     & NVIDIA TITAN-Xp              & -            & 0.8183          & 0.9529          & 0.9752          & -            & -               & 0.9610          & \textbf{0.9781} \\
\textbf{M2U-Net (ours)}                             & NVIDIA GTX 1080 Ti           & 6ms          & 0.8091          & \textbf{0.9630} & 0.9714          & \textbf{7ms} & \textbf{0.8006} & \textbf{0.9703} & 0.9666          \\
\textbf{M2U-Net (ours)}                             & Rockchip RK3399\textsuperscript{*} & 577ms  & 0.8091          & \textbf{0.9630} & 0.9714          & 1.67s        & \textbf{0.8006} & \textbf{0.9703} & 0.9666          \\ \hline
\multicolumn{10}{l}{* Embedded/mobile platform}  \\ \bottomrule
\end{tabular}
\end{center}
\caption{Inference time, Dice score, accuracy, AuC and hardware platform of our proposed method and previous works on DRIVE and CHASE\_DB1.}
\label{tab:previouswork}
\end{table*}
In this section we present the results of M2U-Net on DRIVE, CHASE\_DB1 and HRF in comparison to the performance of U-Net and ERFNet, where possible. 
All networks were trained with the same training setups as M2U-Net and their inference time evaluated on the same hardware (Rockchip RK3399 SoC) and software (PyTorch~\cite{noauthor_pytorch:_2018} with NNPACK~\cite{dukhan_acceleration_2018} backend and TVM~\cite{chen_tvm:_2018}). 
Additionally we discuss the achieved results with regard to previous published literature. 
\subsection{Experimental measurements}
Table~\ref{tab:resultsfourmodels1} shows the model sizes in terms of number of trainable parameters and Multiply-Add operations together with the achieved Dice score. 
For ERFNet, no ARM TVM inference times are reported since the PyTorch implementation as provided by its authors is incompatible with TVM. 

\begin{figure}
\begin{center}
\includegraphics[width=0.50\textwidth]{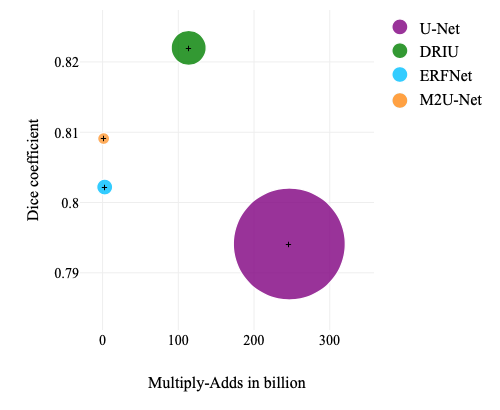}
\end{center}
   \caption{Illustration of model size, performance and computational requirements on DRIVE. The circle sizes are indicative of the number of parameters. 
   	DRIU is the model by Maninis \etal~\cite{maninis_deep_2016}. 
   	Best viewed in color.}
\label{fig:bubble}
\end{figure}

\begin{figure*}
\centering
DRIVE
\begin{multicols}{2}
    \includegraphics[width=\linewidth]{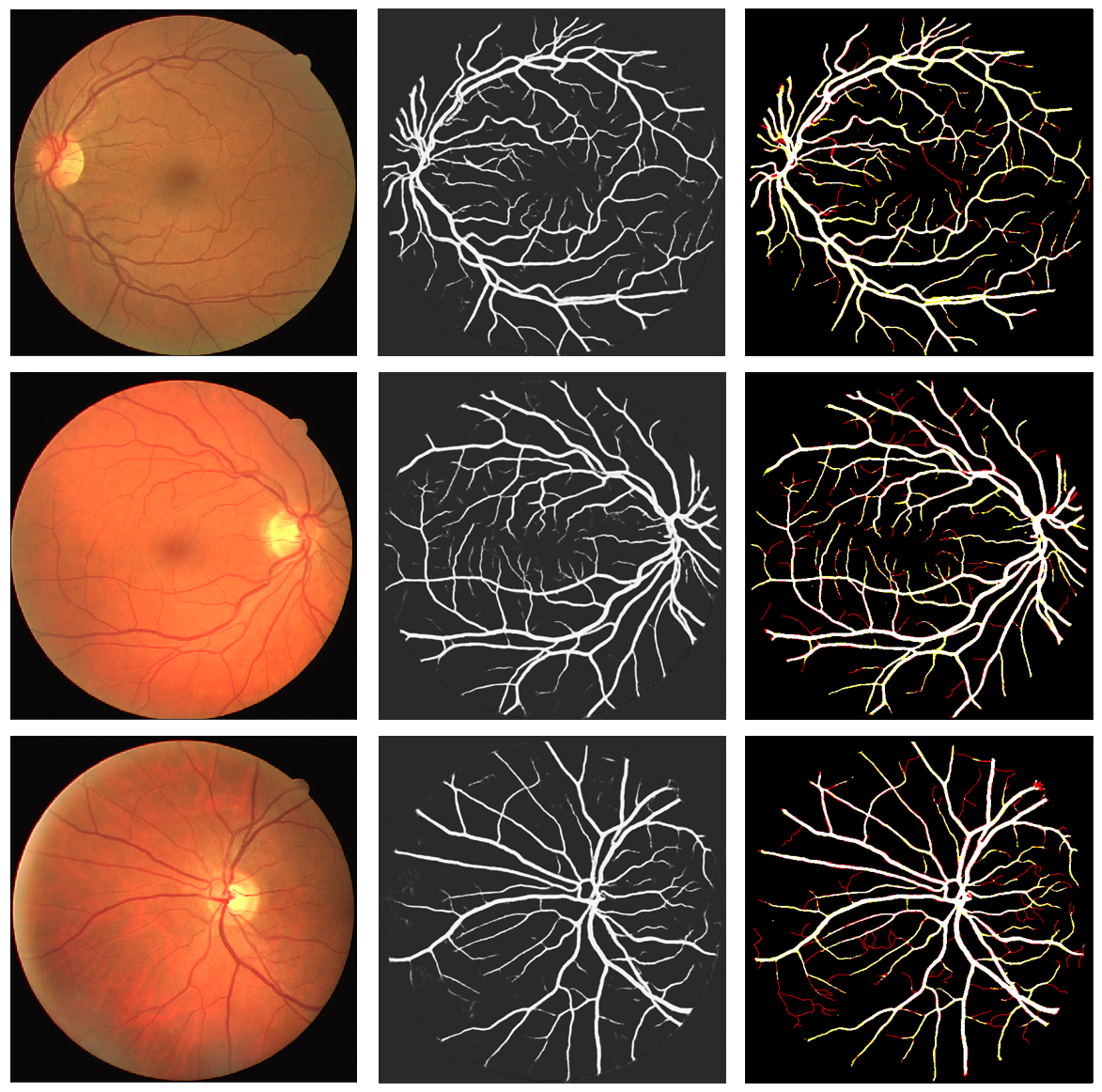}\par 
    \includegraphics[width=\linewidth]{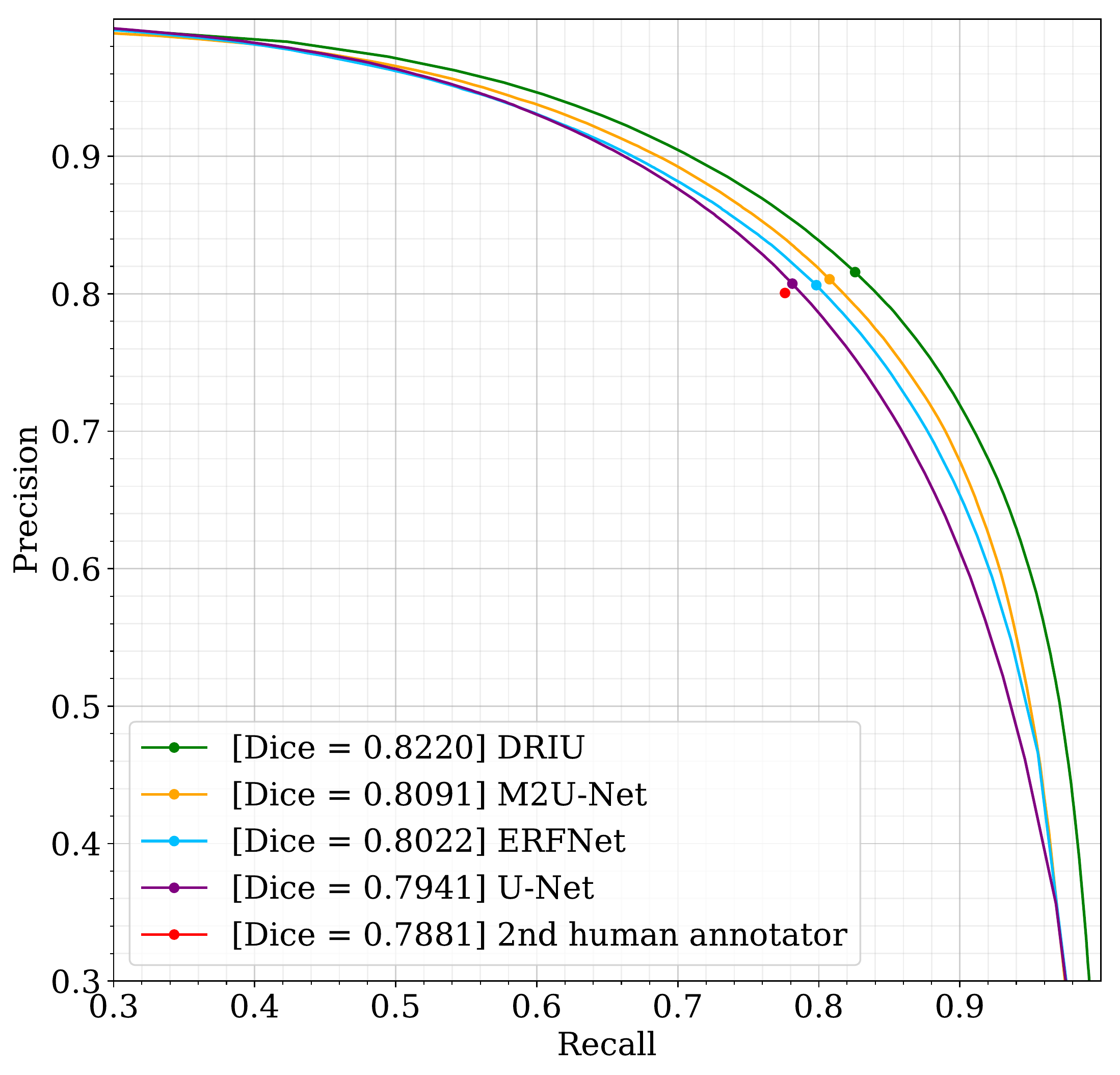}\par 
    \end{multicols}
CHASE\_DB1
\begin{multicols}{2}
    \includegraphics[width=\linewidth]{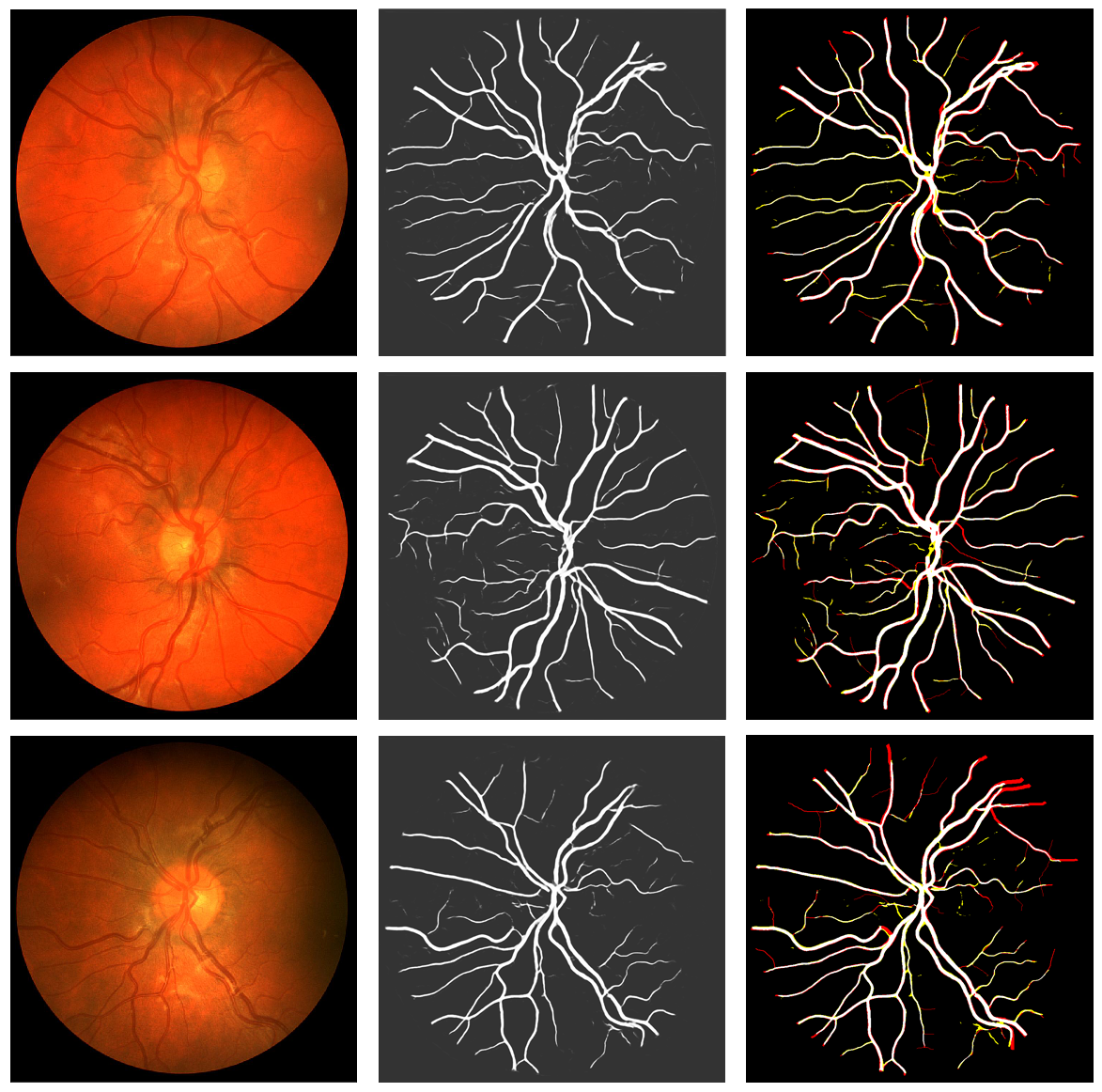}\par
    \includegraphics[width=\linewidth]{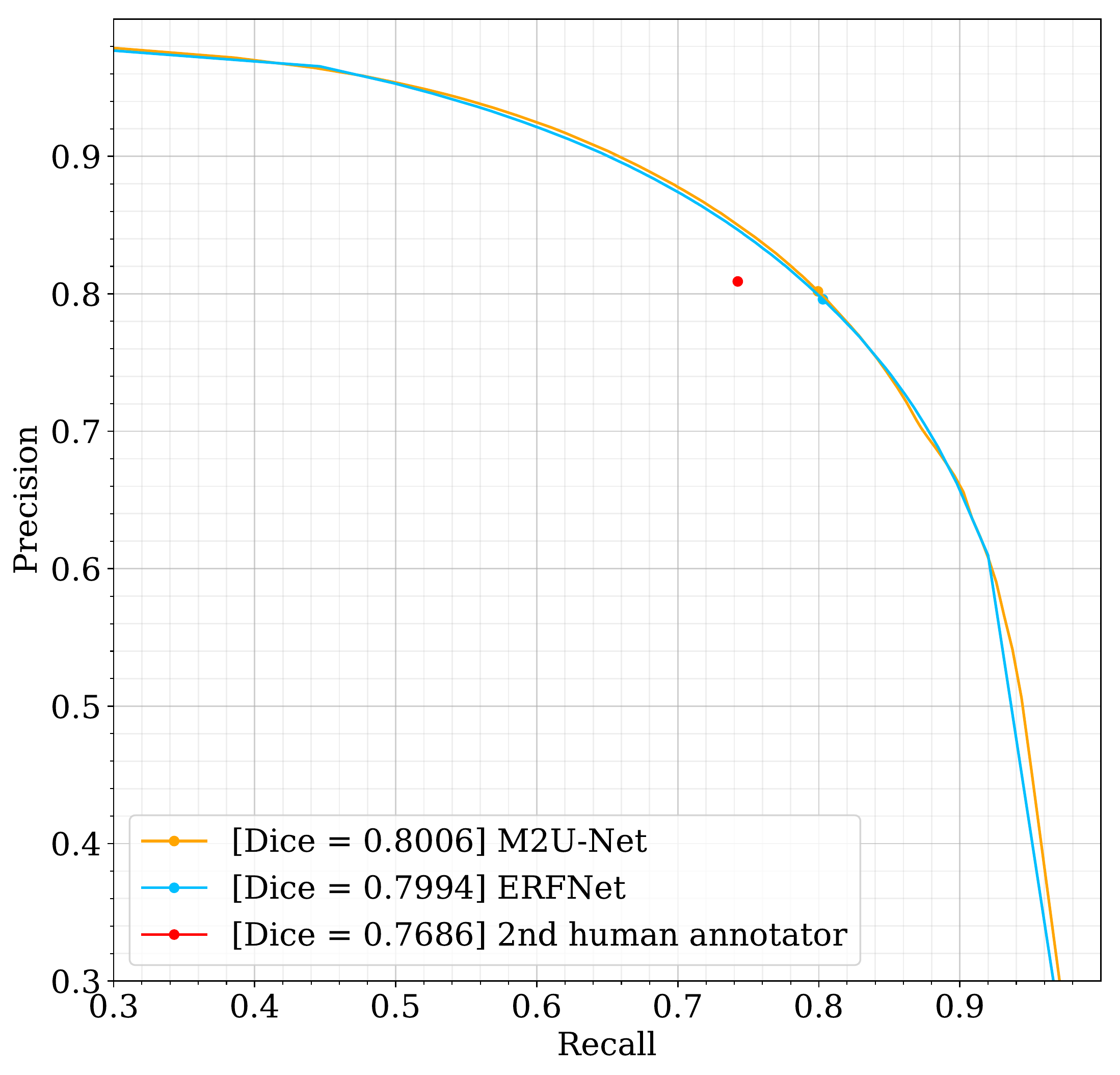}\par
\end{multicols}
\caption{Visualization of results for DRIVE (top) and CHASE\_DB1 (bottom). \textbf{Left}: first column, fundus test image; second column, vessel probability map output of M2U-Net; third column, thresholded binary map with indication of true positives (white), false positives (yellow) and false negatives (red). \textbf{Right}: precision vs. recall curves together with the Dice score at the optimal point along the curve. Best viewed in color.}
\label{fig:resultfig}
\vspace{30pt}
\end{figure*}

Out of the three evaluated networks, M2U-Net is by far the smallest and computationally lightest network with only 0.55M parameters, a size on disk of just 2.2MB and 1.4B multiply-adds for an input of dimension $3 \times 544 \times 544$, as illustarted in Figure~\ref{fig:bubble}.
On DRIVE, our PyTorch implementation has an execution time of under 6 seconds on a single Cortex-A72 compared to over a minute for ERFNet and over 2 hours for U-Net. 
At the same time, the Dice score of M2U-Net on CHASE\_DB1 achieves the best Dice score in our experiment.
Visualizations of the segmented vessel probability map and thresholded binary map of M2U-Net together with precision vs. recall curves for all three networks are shown in Figure~\ref{fig:resultfig}. 

\subsection{Comparison with previous work}
Table~\ref{tab:previouswork} lists previous works in retinal vessel segmentation on DRIVE and CHASE\_DB1 in terms of reported inference time, hardware platform, Dice score, accuracy, and AuC. 
On CHASE\_DB1, our proposed architecture achieves a new state-of-the-art Dice score of 0.8006 and an inference time of 1.67s on a Rockchip RK3399 SoC and 7ms on a NVIDIA GTX 1080Ti. In contrast, the second fastest method~\cite{orlando_discriminatively_2017} takes 2.7s on a Intel Xeon CPU with a Dice score of 0.7332.
On DRIVE our method achieves a Dice score that is 1.3 percentage points lower compared to~\cite{maninis_deep_2016}, but has $14\times$ less parameters.

On HRF, M2U-Net achieves new state-of-the-art results in terms of Dice score and inference times as shown in Table~\ref{tab:previousworkhrf}. 
Compared to the previous best unsupervised and supervised method we report an improvement in Dice score of 2.36 and 6.02 percentage points respectively. 
On ARM M2U-Net achieves inference times of 14.7s and on a GPU it achieves inference speed of 19.9ms, making it the first real-time method on this high-resolution dataset ($2336 \times 3504$). 

The method by \cite{maninis_deep_2016} might be able to achieve higher quality results on HRF and CHASE\_DB1 compared to ours, however due to it's larger size, we expect it to face memory constraints when trained with full resolution images.

The power consumption measured during inference of our method on a Rockchip RK3399 SoC development board is 9.6W with an idle power consumption of 3.6W. 
This is substantially lower than the power consumption of common server grade GPUs that is often above 200W. 
The only other work in the reviewed literature that explicitly reports power consumption measurements \cite{koukounis_high_2014}, reports a power consumption of 2.89W and takes 52.3ms on a Spartan 6 FPGA, which is not directly comparable. Optimized FPGA implementations are generally faster and more energy efficient than ARM implementations, and we would expect similar gains for our model on FPGA. The model of \cite{koukounis_high_2014}, however achieves an accuracy of only 0.9240 and an AuC of 0.9008 on DRIVE, compared to an accuracy of 0.9630 and AuC of 0.9714 of our method (no Dice score is reported in~\cite{koukounis_high_2014}).

\begin{table}[]
\small
\begin{tabular}{ccccc}
\hline
Method                                                & Plt.                                  & Time            & Dice            & Acc                          \\ \hline
\textbf{Unsupervised}                                 &                                       &                 &                 &                               \\ \hline
Annunziata \etal~\cite{annunziata_leveraging_2016}    & CPU\textsuperscript{*}                                   & -               & 0.7578          & 0.9581                         \\
Budai \etal~\cite{budai_robust_2013}                  & CPU\textsuperscript{\textdagger}                & 26.7s           & -               & 0.9610                         \\
Odstrcilik \etal~\cite{odstrcilik_retinal_2013}       & CPU\textsuperscript{\textdaggerdbl}      & 92s             & 0.7324          & 0.9494                         \\

Zhang \etal~\cite{zhang_robust_2016}                 & -                                     & -               & -               & 0.9556          \\
\textbf{Supervised}                                   &                                       &                 &                 &                 \\  \hline
Orlando \etal~\cite{orlando_discriminatively_2017}    & CPU\textsuperscript{\textparagraph}   & 5.8s            & 0.7158          & -                              \\
Yan \etal~\cite{yan_joint_2018}                       & GPU\textsuperscript{\textsection}   & -               & 0.7212          & 0.9437                    \\
\textbf{M2U-Net (ours)}                               & GPU\textsuperscript{\textbar\textbar}     & \textbf{19.9ms} & \textbf{0.7814} & \textbf{0.9635}  \\
\textbf{M2U-Net (ours)}                               & ARM\textsuperscript{\#} & 14.7s           & \textbf{0.7814} & \textbf{0.9635}  \\ \hline
\multicolumn{5}{l}{* AMD A4-3300M} \\
\multicolumn{5}{l}{\textdagger{} 2.3 GHz }  \\
\multicolumn{5}{l}{\textdaggerdbl{} Intel Core i7}  \\
\multicolumn{5}{l}{\textparagraph{} Intel Xeon E5-2690}  \\
\multicolumn{5}{l}{\textsection{} NVIDIA TITAN-Xp}  \\ 
\multicolumn{5}{l}{\textbar\textbar{} NVIDIA GTX 1080 Ti} \\  
\multicolumn{5}{l}{\# Rockchip RK3399 (TVM)} \\\bottomrule
\end{tabular}
\caption{Inference time, Dice score, accuracy and hardware platform of our proposed method and previous works on HRF.}
\label{tab:previousworkhrf}
\end{table}

\subsection{Ablation study}
In our experiments with various values for $t$ in the decoder block, we found that values smaller than 1 result in similar Dice scores to values greater than 1, while at the same time substantially reducing the parameter count and memory footprint of the network. 
We also tried using only depthwise separable convolutions after each upsampling and concatenate operation instead of Stride=1 bottleneck blocks and found slightly worse performance in terms of Dice score while having a higher number of parameters. 

\section{Conclusion}
We introduced M2U-Net, a new light-weight neural network architecture for the segmentation of retinal vasculature with novel contracting bottleneck blocks in the decoder part and elements of a pretrained MobileNetV2 in the encoder part.

Furthermore, it will enable new practical applications, where high quality analysis of high resolution images in real-time is required, e.g. in robotic microsurgery of the eye. 
We expect that the architecture can also be successfully applied to related tasks, such as road segmentation on satellite images or the semantic segmentation of urban scenes.

\section*{Acknowledgements}
This research is carried out in collaboration with The London Project to Cure Blindness, ORBIT, Institute of Ophthalmology, University College London (UCL), London, UK., NIHR Biomedical Research Centre at Moorfields Eye Hospital NHS Foundation Trust, UCL Institute of Ophthalmology, London, UK., Moorfields Eye Hospital NHS Foundation Trust, London, UK. and is partly funded by The Sir Michael Uren Foundation R170010A.
{\small
\bibliographystyle{ieee}
\bibliography{m2unet}
}

\end{document}